\documentclass{article}

\usepackage[preprint]{neurips_2026}
\usepackage{placeins}

\usepackage[utf8]{inputenc} 
\usepackage[T1]{fontenc}    
\usepackage{hyperref}       
\usepackage{url}            
\usepackage{booktabs}       
\usepackage{amsfonts}       
\usepackage{nicefrac}       
\usepackage{microtype}      
\usepackage{xcolor}         

\usepackage{tikz}
\usepackage{amsmath,amssymb}
\usetikzlibrary{shapes.geometric, arrows.meta, positioning, fit, backgrounds, calc}
 
\definecolor{goldline}{HTML}{A8842C}
\definecolor{goldfill}{HTML}{FBF1DC}
\definecolor{goldtext}{HTML}{8C6D1F}
\definecolor{graybox}{HTML}{E7E7E7}
\definecolor{graytext}{HTML}{555555}
\definecolor{darktext}{HTML}{1A1A1A}

\title{Probing Latent Colombian Identity Inferences in Qwen2.5-7B with Natural Language Autoencoders}

\author{%
  Pablo Santiago Potes Velasco \\
  Universidad Autónoma de Occidente \\
  Cali, Colombia \\
  \texttt{pablo.potes@uao.edu.co} \\
  \And
  María del Mar García Matabanchoy \\
  Universidad Autónoma de Occidente \\
  Cali, Colombia \\
  \texttt{maria\_d.garcia\_m@uao.edu.co} \\
  \And
  Óscar Julián Pérez Ladino \\
  Universidad Autónoma de Occidente \\
  Cali, Colombia \\
  \texttt{oscar.perez@uao.edu.co} \\
  \And
  Jhoan Stevan Mosquera Ortiz \\
  Universidad Autónoma de Occidente \\
  Cali, Colombia \\
  \texttt{jhoan.mosquera@uao.edu.co} \\
  \And
  Nicolás Lozano Mazuera \\
  Universidad Autónoma de Occidente \\
  Cali, Colombia \\
  \texttt{nicolas.lozano\_m@uao.edu.co} \\
  \And
  Gilber Alexis Corrales Gallego \\
  Universidad Autónoma de Occidente \\
  GobLab-UAI \\
  Cali, Colombia \\
  \texttt{gacorrales@uao.edu.co}
}

\begin{document}

\maketitle

\begin{abstract}
Large language models may infer demographic attributes from subtle linguistic cues even when those attributes are not explicitly stated. This pilot study examines whether Qwen2.5-7B-Instruct internally represents Colombian identity, socioeconomic status, or stereotype-related information when processing Colombian-Spanish and English prompts. We use Natural Language Autoencoders (NLA) to verbalize residual-stream activations from layer 20 across four positional quartiles per prompt. Our dataset contains 30 prompts arranged as 15 matched Spanish-English pairs, spanning explicit Colombian cues, implicit Colombian cues, and neutral controls. We report descriptive rates and qualitative evidence rather than statistically powered effects, focusing on whether latent nationality or stereotype representations appear before they are verbalized in the model output. This work connects activation-level interpretability with bias evaluation for underrepresented Spanish varieties.
\end{abstract}

\section{Introduction}
\label{sec:introduction}

Large language models infer user attributes—nationality, dialect, socioeconomic background—from cues never stated explicitly, and these inferences shape generation even when unverbalized: instruction-tuned models refuse over 98\% of explicit demographic queries yet still condition outputs on inferred attributes \citep{Bouchaud2025Linear}. \citet{FraserTaliente2026NLA} illustrate this directly with Natural Language Autoencoders (NLA): their \emph{language-switching} case study shows a model representing a user as Russian several tokens before any lexical cue, later traced to mislabeled training data. Whether an analogous unverbalized inference occurs for Spanish is unknown, despite evidence it should matter most for the varieties most often misread—Colombian Spanish scores far below Peninsular Spanish in recognition benchmarks (F1$=0.282$ vs.\ $0.723$) for reasons tracking training-data composition \citep{Kawasaki2026Digital, Mayor2025It}, yet no study inspects what the model represents internally before producing that output.
 
We use the open-source NLA for Qwen-2.5-7B on $\sim$30 matched Colombian-Spanish/English prompt pairs to probe for latent nationality, socioeconomic, or stereotype representations absent from the model's final response. Section~\ref{sec:related-work} situates this work; Section~\ref{sec:method} details our procedure; Section~\ref{sec:results} reports findings.

\section{Related Work}
\label{sec:related-work}

Interpreting what a language model represents about its input without supervised probes has converged on two strategies: projecting or patching activations to recover output-relevant information without training \citep{Nostalgebraist2020LogitLens, Belrose2023TunedLens, Ghandeharioun2024Patchscopes}, and training a reader model—via sparse dictionaries \citep{Cunningham2024SAE} or a full verbalizer—to translate activations into free text \citep{Chen2024SelfIE, Pan2024LatentQA, Karvonen2025ActivationOracles}. We adopt the latter, using the Natural Language Autoencoder (NLA) of \citet{FraserTaliente2026NLA} as our probing instrument; their \emph{language-switching} case study, where the verbalizer surfaces a latent nationality inference before any lexical cue appears in the prompt, motivates asking whether an analogous unverbalized inference occurs for Colombian Spanish.
 
A parallel line indicates that demographic attributes are linearly decodable from activations, regardless of the verbalization method. \citet{Bouchaud2025Linear} report AUC-ROC up to 0.995 for probing gender, race, and socioeconomic status from indirect cues, concentrated in middle layers; \citet{Lauscher2022SocioProbe} and \citet{Tang2023What} corroborate this across architectures, while \citet{Hu2026Race} shows finer-grained attributes are distributed rather than strictly linear. Most relevant to our hypothesis is the \emph{alignment gap}: instruction-tuned models refuse over 98\% of explicit demographic queries \citep{Bouchaud2025Linear} yet still condition generation on stereotype-aligned inferences when the attribute is only implied \citep{Neplenbroek2025Reading, Tang2023What}, consistent with bias persisting in contextualized representations after debiasing \citep{Guo2021Detecting, Tan2019Assessing, Bommasani2020Interpreting, Huang2020Reducing, Zhang2025Semantic}.
 
Behaviorally, this latent inference produces measurable disparities for Spanish varieties and non-native English speakers. Peninsular Spanish is consistently best recognized and generated, while Latin American varieties lag substantially—Colombia among the lowest (F1$=0.282$ vs.\ $0.723$ for Spain)—tracking training-data composition rather than digital resource volume \citep{Kawasaki2026Digital, Mayor2025It, Martinez2025Spanish}, and English bias-mitigation techniques do not transfer to Spanish \citep{Robles2025SESGO}. English shows analogous effects: anchoring on perceived non-nativeness degrades response quality \citep{Reusens2024Native}, minoritized dialects receive more stereotyped and condescending outputs \citep{Fleisig2024Linguistic}, and alignment training widens rather than narrows these gaps \citep{Ryan2024Unintended, Mire2025Rejected, Nayeem2026Which, Kantharuban2023Quantifying}.
 
These two literatures remain unconnected: linear-probe studies establish \emph{that} demographic inference diverges from verbalized output, and bias-audit studies establish \emph{that} this divergence harms underrepresented varieties, but none use a training-free, unsupervised verbalizer with controlled explicit/implicit/neutral elicitation to localize \emph{when} a Colombian-identity inference emerges from indirect cues within the prompt itself. This is the gap we target.

\section{Method}
\label{sec:method}
\paragraph{Overview.}
\citep{FraserTaliente2026NLA} show that Natural Language Autoencoder (NLA) explanations can reveal that a model internally represents a user as Russian \emph{before} any unambiguous lexical cue and \emph{before} this belief is verbalized. We adapt this finding into a controlled probe of whether Qwen2.5-7B-Instruct internally infers \emph{Colombian} identity from a single implicit cue, using the open-source NLA pair released with~\citep{FraserTaliente2026NLA}.
 
\paragraph{Extraction and quartile sampling.}
Let $M$ be the target model and $h_\ell \in \mathbb{R}^d$ ($d{=}3584$) the residual-stream activation at layer $\ell{=}20$, the depth at which the released Qwen2.5-7B NLA pair was trained, and an $N$-token prompt. A single deterministic forward pass of $M$ yields the full sequence of activations $\{h_\ell[1],\dots,h_\ell[N]\}$, one per token position. From this sequence we select four activations, the \emph{last} token of each of four contiguous quartiles,
\[
q_k = \lfloor k N / 4 \rfloor, \quad k \in \{1,2,3,4\},
\]
so $h_\ell[q_1],\dots,h_\ell[q_4]$ are four distinct, full-dimensional vectors drawn from four different positions in the same forward pass — not a single activation, and not a partition of $h_\ell$'s $3584$ dimensions. Querying the AV at every position is infeasible, since each call emits hundreds of tokens, so this four-point subsample keeps the per-prompt AV cost fixed regardless of $N$~\citep{FraserTaliente2026NLA}. The last-token choice additionally keeps each query in-distribution: the AV's supervised warm-start was trained only on prefix-final activations, and each $q_k$ is, formally, the final token of prefix $x[1{:}q_k]$. Each $h_\ell[q_k]$ is passed once (no resampling) to the AV, $z_k \sim \mathrm{AV}(\cdot \mid h_\ell[q_k])$, yielding one explanation per quartile.

\paragraph{Design.}
We construct 15 base scenarios, each realized in Spanish and in English translation (30 prompts), with five scenarios per \emph{explicitness} level: \textbf{explicit} (unambiguous Colombian marker, positive control), \textbf{implicit} (exactly one subtle cue, analogous to the \emph{vodka}$\to$``Russian'' case in~\citep{FraserTaliente2026NLA}), and \textbf{neutral} (topically matched, no national marker, negative control distinguishing a Colombia-specific effect from generic AV confabulation).
 
\begin{table}[h]
\centering
\small
\begin{tabular}{lccc}
\toprule
 & Explicit ($n{=}5$) & Implicit ($n{=}5$) & Neutral ($n{=}5$) \\
\midrule
Spanish & explicit cue & single implicit cue & no cue \\
English & explicit cue & single implicit cue & no cue \\
\bottomrule
\end{tabular}
\caption{2 (language) $\times$ 3 (explicitness) design, $n{=}30$ prompts.}
\end{table}
 
\paragraph{Structured coding.}
\label{par:coding}
Each prompt's four quartile-level explanations $(z_1,z_2,z_3,z_4)$ are expanded into $120$ prompt$\times$quartile units and independently coded by Claude Sonnet for \emph{nationality}, \emph{socioeconomic status}, and \emph{stereotype} mentions. The coding prompt requires (i) a category marked \texttt{true} only on explicit textual evidence, (ii) default \texttt{false} for explanations consisting solely of generic model self-description (e.g.\ \emph{``I am Qwen, a large language model created by Alibaba Cloud''}), a known AV failure mode rather than a substantive judgment about the prompt, and (iii) a verbatim supporting quote for every positive label. An automated audit confirms full coverage, flags failed API calls, and rejects any returned quote not found \emph{verbatim} in its source explanation; one of $120$ units failed extraction after retries and is excluded listwise ($n{=}119$).
 
Because the AV is known to confabulate plausible-sounding but contextually wrong specifics — in our data, substituting Spain, Turkey, or Canada for Colombia in an otherwise on-theme explanation — a positive nationality label conflates two distinct events: representing \emph{some} national identity versus \emph{Colombian} identity specifically. We therefore additionally tag each positive nationality citation for whether it names Colombia (or an unambiguous metonym, e.g.\ Bogotá, TransMilenio), and report both the unrestricted and the Colombia-specific rate in \S\ref{sec:results}; this distinction is fixed at analysis design time, not introduced post hoc.
 
\paragraph{Statistical analysis.}
\label{par:stats}
We report each (group $\times$ quartile) mention rate with a Wilson score interval, valid at this design's small per-cell counts ($n{=}9$--$10$) where a normal approximation would not be. Implicit-vs-neutral differences at each quartile are tested with Fisher's exact test on the corresponding $2{\times}2$ table, rather than a $\chi^2$ test whose asymptotic assumptions are unmet here. To summarize \emph{when} nationality is first represented, we take, for each prompt where the category appears at least once, its earliest quartile, and compare distributions between groups with a Mann–Whitney $U$ test; prompts where it never appears are reported as a separate per-group count rather than imputed, since their absence is itself informative. All tests are two-sided and uncorrected for multiple comparisons, consistent with the exploratory scope below.
 
\paragraph{Scope.} With $n{=}5$ per cell, results are reported as descriptive rates with qualitative evidence rather than as statistically powered effects; we extract from a single layer and a single AV sample per quartile, given hackathon-scale compute.

\section{Results}
\label{sec:results}

\begin{figure}[h]
\centering
\includegraphics[width=0.7\linewidth]{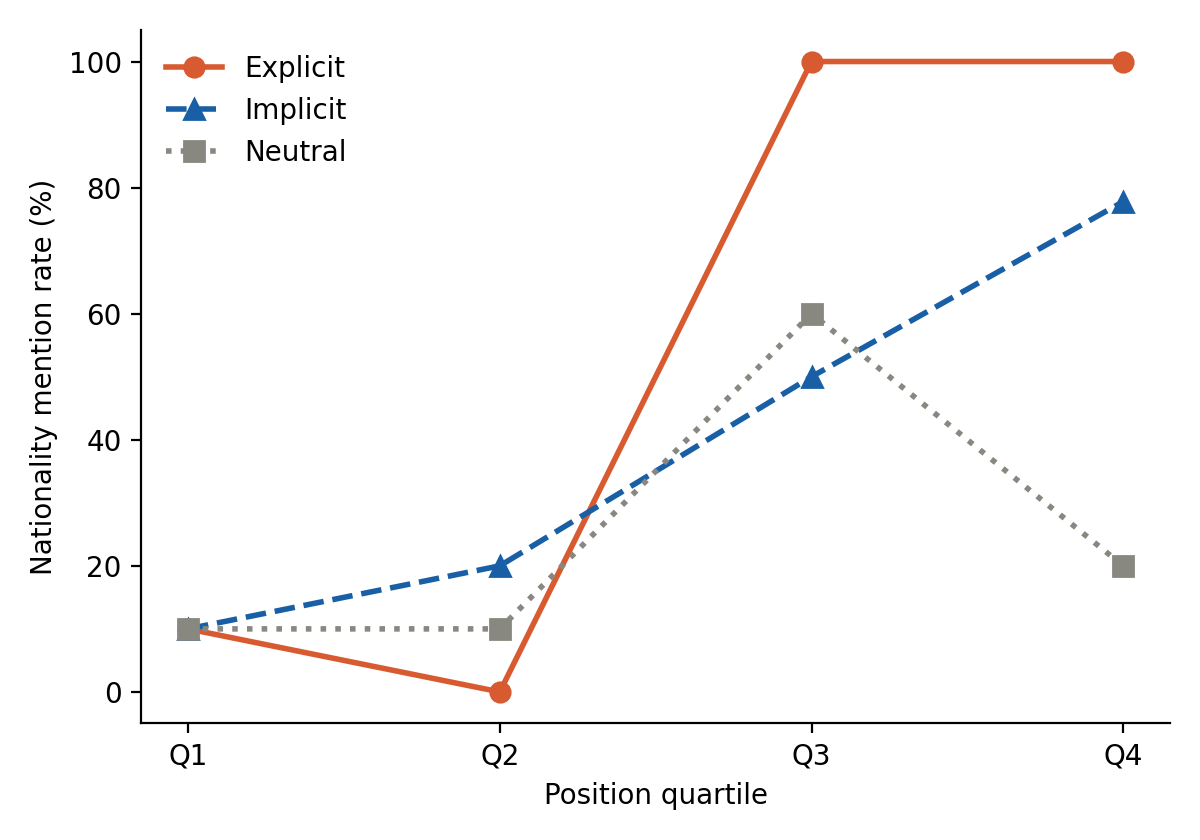}
\caption{Nationality mention rate by quartile, all three groups. Implicit (blue, triangles) rises steadily from Q1 to Q4; explicit (orange, circles) ceils early; neutral (gray, squares) is non-monotonic, addressed in \S\ref{sec:results}.}
\label{fig:curva_cuartiles}
\end{figure}

Table~\ref{tab:tasas_cuartil} reports the rate at which AV explanations mention nationality, for each group and quartile. The implicit-cue group rises monotonically across all four quartiles (0.10$\to$0.78), consistent with the context-accumulation hypothesis; Figure~\ref{fig:curva_cuartiles} shows this as a sustained positive slope with neither the early ceiling of the explicit control nor a late collapse. Two cells deviate from monotonicity — explicit at Q2 (0.00) and neutral at Q4 (0.20) — and we resolve both below before turning to the quartile of first appearance (Table~\ref{tab:primera_aparicion}) and the implicit-vs-neutral comparison (Table~\ref{tab:fisher}).

\begin{table}[h]
\centering
\small
\begin{tabular}{lcccc}
\toprule
 & Q1 & Q2 & Q3 & Q4 \\
\midrule
Explicit & 0.10 & 0.00 & 1.00 & 1.00 \\
Implicit & 0.10 & 0.20 & 0.50 & 0.78 \\
Neutral  & 0.10 & 0.10 & 0.60 & 0.20 \\
\bottomrule
\end{tabular}
\caption{Nationality mention rate by group and quartile ($n{=}9$--$10$ per cell; one unit excluded for failed extraction, \S\ref{par:coding}).}
\label{tab:tasas_cuartil}
\end{table}

\paragraph{Both irregularities trace to non-Colombian confabulation, not noise.} We inspected the source quotes behind explicit/Q2 and neutral/Q4 directly. Every Q2 explanation in the explicit group is off-theme confabulation unrelated to the prompt (e.g.\ \emph{``Marketing de Google''}, \emph{``conferencia Tesla''}) — the same low-anchoring failure mode documented at Q1 (\S\ref{par:coding}), recurring at a second early position for short prompts. The two positive citations behind neutral/Q4 name Spain and paella, never Colombia. Restricting the nationality variable to Colombia-specific mentions (Spain/Turkey/Canada excluded; \S\ref{par:coding}) removes both irregularities: neutral falls to $0.00$ at every quartile, while implicit remains strictly above zero from Q2 onward ($0.00/0.10/0.20/0.11$). The qualitative pattern in Figure~\ref{fig:curva_cuartiles} therefore understates, rather than fabricates, the implicit-vs-neutral separation; we report the unrestricted rate in Table~\ref{tab:tasas_cuartil} for comparability with the AV's overall behavior, and the Colombia-specific rate as the variable of record for testing the hypothesis.

\begin{table}[h]
\centering
\small
\begin{tabular}{lccc}
\toprule
 & Mean Q of 1st mention & Median & Never mention ($n$) \\
\midrule
Explicit & 2.80 & 3.0 & 0/10 \\
Implicit & 2.88 & 3.0 & 2/10 \\
Neutral  & 2.50 & 3.0 & 4/10 \\
\bottomrule
\end{tabular}
\caption{Quartile of first nationality mention, computed over prompts where the category appears at least once; the right column reports prompts where it never appears.}
\label{tab:primera_aparicion}
\end{table}

Table~\ref{tab:primera_aparicion} shows comparable mean onset quartiles across groups ($\approx$Q3) among prompts that ever trigger a mention, but a markedly different \emph{rate} of never triggering one at all: $0/10$ for explicit, $2/10$ for implicit, $4/10$ for neutral. This count — not the onset quartile — carries most of the between-group signal, and is reported alongside the mean rather than imputed into it.

\begin{table}[h]
\centering
\small
\begin{tabular}{lccccc}
\toprule
Quartile & $k$ implicit & $n$ implicit & $k$ neutral & $n$ neutral & $p$ (Fisher) \\
\midrule
Q1 & 1 & 10 & 1 & 10 & 1.000 \\
Q2 & 2 & 10 & 1 & 10 & 1.000 \\
Q3 & 5 & 10 & 6 & 10 & 1.000 \\
Q4 & 7 & 9  & 2 & 10 & \textbf{0.023} \\
\bottomrule
\end{tabular}
\caption{Fisher's exact test, implicit vs.\ neutral, by quartile.}
\label{tab:fisher}
\end{table}

Only Q4 reaches conventional significance ($p{=}0.023$); given $n{\le}10$ per cell and four uncorrected comparisons, we read this as directional evidence for a late-quartile separation rather than a confirmatory effect (\S\ref{par:stats}), consistent with the exploratory scope of this study.

See Appendix~\ref{sec:case-studies} for a qualitative, case-by-case analysis of individual prompts.
 
\section{Conclusion}
\label{sec:conclusion}
 
Using an unsupervised, training-free verbalizer, we find that Qwen2.5-7B-Instruct's residual stream comes to represent Colombian identity from a single implicit lexical cue, with the Colombia-specific mention rate rising from $0.00$ at Q1 to $0.20$ at Q3 while the neutral control remains at $0.00$ throughout. This separation reaches conventional significance only at the final quartile ($p{=}0.023$) and rests on $n{=}5$ scenarios per cell; we report it as directional evidence that an unverbalized nationality inference can emerge from a single cue, not as a confirmed effect. Two apparent irregularities in the raw mention rate were traced to a known AV failure mode — confabulating a contextually wrong country — and resolved by restricting to Colombia-specific citations, a check we recommend for any reuse of NLA explanations as a regional-identity probe. The natural next step is repeating this design at higher $n$ per cell and at a second residual-stream layer, to test whether the late-quartile separation we observe is a property of this specific depth or holds more generally.

\bibliographystyle{plainnat}
\bibliography{ref}

\appendix

\section{Technical appendices and supplementary material}~\label{sec:case-studies}

\begin{figure}[htbp]
\centering
\resizebox{0.8\columnwidth}{!}{%
\begin{tikzpicture}[
    font=\sffamily,
    every node/.style={font=\sffamily},
    leafbox/.style={
      draw=goldline, thick, fill=goldfill, rounded corners=2pt,
      minimum width=3.6cm, minimum height=0.8cm, align=center,
      text=darktext, font=\sffamily\small
    },
    headbox/.style={
      draw=goldline, line width=1pt, fill=goldline, rounded corners=2pt,
      minimum width=3.6cm, minimum height=0.9cm, align=center,
      text=white, font=\sffamily\bfseries\small
    },
    rootbox/.style={
      draw=darktext, thick, fill=graybox, rounded corners=2pt,
      minimum width=4.4cm, minimum height=1cm, align=center,
      text=darktext, font=\sffamily\bfseries
    },
    branch/.style={-{Stealth[length=2.2mm]}, goldline, thick},
]
 
\node[rootbox] (root) at (0,5.4)
  {30 prompts $=$ 15 base stories $\times$ 2 languages};
 
\node[headbox] (es) at (-4.2,3.7) {Spanish (ES) \quad $n{=}15$};
\node[headbox] (en) at ( 4.2,3.7) {English (EN, translation) \quad $n{=}15$};
 
\draw[branch] (root.south) -- ++(0,-0.35) -| (es.north);
\draw[branch] (root.south) -- ++(0,-0.35) -| (en.north);
 
\node[leafbox] (es1) at (-4.2,2.3) {5 $\times$ Explicit Colombian};
\node[leafbox] (es2) at (-4.2,1.3) {5 $\times$ Implicit Colombian};
\node[leafbox] (es3) at (-4.2,0.3) {5 $\times$ Neutral};
 
\draw[goldline, thick] (es.south) -- (es1.north);
\draw[goldline, thick] (es1.south) -- (es2.north);
\draw[goldline, thick] (es2.south) -- (es3.north);
 
\node[leafbox] (en1) at (4.2,2.3) {5 $\times$ Explicit Colombian};
\node[leafbox] (en2) at (4.2,1.3) {5 $\times$ Implicit Colombian};
\node[leafbox] (en3) at (4.2,0.3) {5 $\times$ Neutral};
 
\draw[goldline, thick] (en.south) -- (en1.north);
\draw[goldline, thick] (en1.south) -- (en2.north);
\draw[goldline, thick] (en2.south) -- (en3.north);
 
\foreach \i/\j in {es1/en1, es2/en2, es3/en3}{
  \draw[graytext, dashed, thin] (\i.east) -- (\j.west)
    node[midway, fill=white, inner sep=1pt, font=\scriptsize\itshape, text=graytext] {pair};
}
\end{tikzpicture}%
}
\caption{\textbf{Dataset structure.} Each of the 15 base stories is realized in Spanish-English pairs; the explicitness factor (explicit / implicit / neutral) varies \emph{between} stories, yielding $n{=}5$ per cell.}
\label{fig:dataset_structure}
\end{figure}
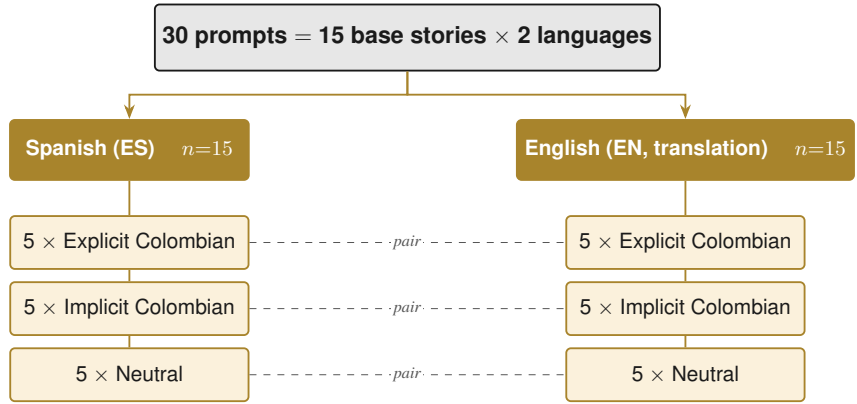

\vspace{0.2cm}

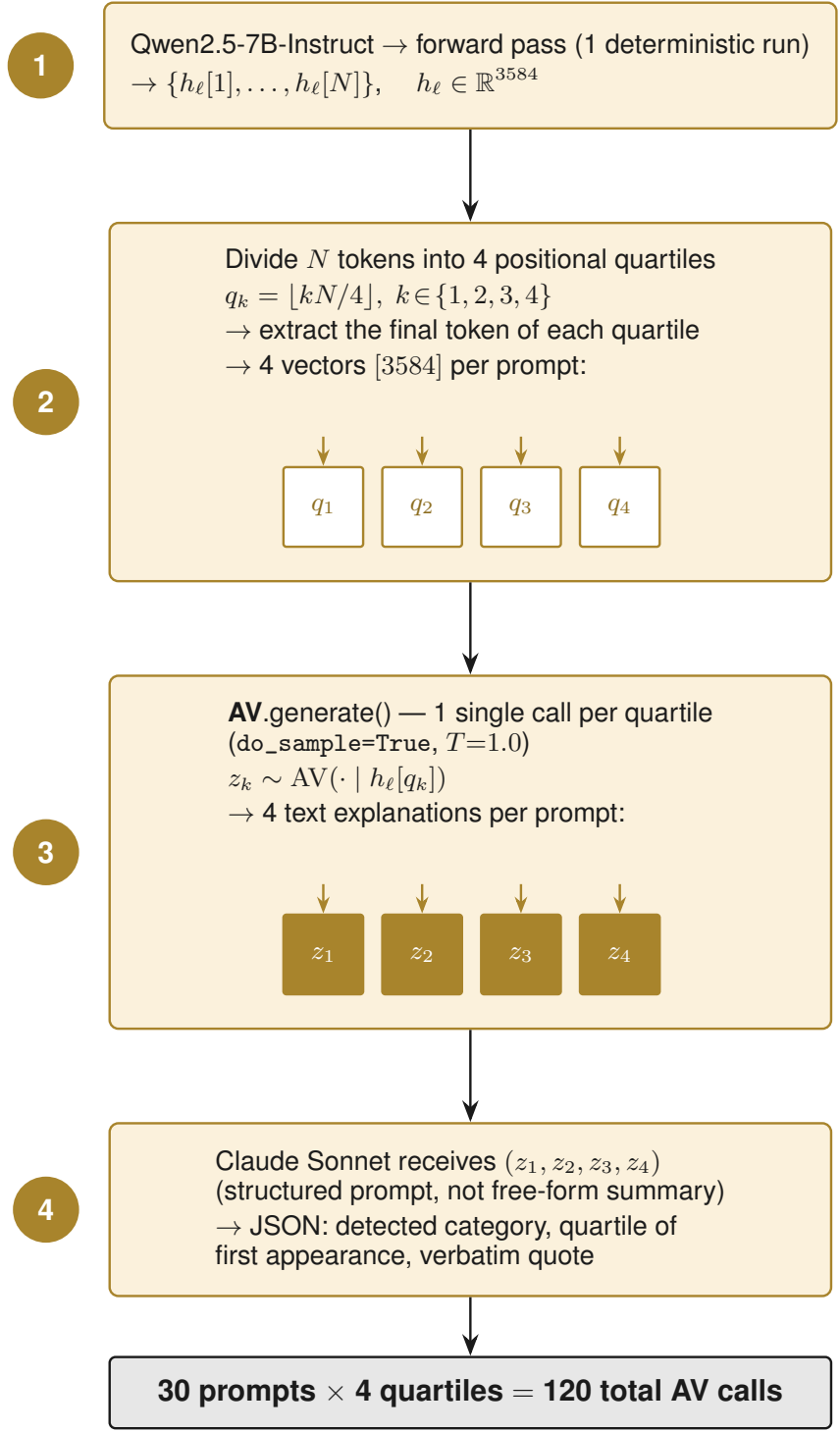
\begin{figure}[htbp]
\centering
\resizebox{0.8\textwidth}{!}{%
\begin{tikzpicture}[
    font=\sffamily,
    stepbadge/.style={
      circle, draw=goldline, line width=1.1pt, fill=goldline,
      minimum size=0.75cm, text=white, font=\sffamily\bfseries
    },
    stepbox/.style={
      draw=goldline, thick, fill=goldfill, rounded corners=3pt,
      minimum width=8.6cm, align=left,
      inner sep=9pt, text=darktext, font=\sffamily\small
    },
    qtbox/.style={
      draw=goldline, thick, fill=white, rounded corners=1pt,
      minimum width=0.95cm, minimum height=0.95cm, align=center,
      text=goldtext, font=\sffamily\bfseries\small
    },
    zbox/.style={
      draw=goldline, thick, fill=goldline, rounded corners=1pt,
      minimum width=0.95cm, minimum height=0.95cm, align=center,
      text=white, font=\sffamily\bfseries\small
    },
    arrowflow/.style={-{Stealth[length=2.6mm]}, darktext, thick},
]
 
\node[stepbox, minimum height=1.5cm] (s1) at (0,0)
  {Qwen2.5-7B-Instruct $\to$ forward pass (1 deterministic run) \\[3pt]
   $\to \{h_\ell[1], \dots, h_\ell[N]\}$, \quad $h_\ell \in \mathbb{R}^{3584}$};
\node[stepbadge, left=0.35cm of s1] {1};
 
\node[stepbox, below=1.1cm of s1] (s2)
  {Divide $N$ tokens into 4 positional quartiles \\[2pt]
   $q_k = \lfloor kN/4 \rfloor,\; k\!\in\!\{1,2,3,4\}$ \\[2pt]
   $\to$ extract the final token of each quartile \\[2pt]
   $\to$ 4 vectors $[3584]$ per prompt: \\[1.8cm]}; 
\node[stepbadge, left=0.35cm of s2] {2};
 
\path (s2.south) ++(0, 0.9) coordinate (q_center);
\node[qtbox] (q2) at ($(q_center) + (-0.575, 0)$) {$q_2$};
\node[qtbox] (q1) [left=0.2cm of q2] {$q_1$};
\node[qtbox] (q3) [right=0.2cm of q2] {$q_3$};
\node[qtbox] (q4) [right=0.2cm of q3] {$q_4$};

\draw[goldline, thick, -{Stealth[length=2mm]}] ($(q1.north) + (0, 0.35)$) -- (q1.north);
\draw[goldline, thick, -{Stealth[length=2mm]}] ($(q2.north) + (0, 0.35)$) -- (q2.north);
\draw[goldline, thick, -{Stealth[length=2mm]}] ($(q3.north) + (0, 0.35)$) -- (q3.north);
\draw[goldline, thick, -{Stealth[length=2mm]}] ($(q4.north) + (0, 0.35)$) -- (q4.north);
 
\node[stepbox, below=1.1cm of s2] (s3)
  {\textbf{AV}.generate() --- 1 single call per quartile \\
   (\texttt{do\_sample=True}, $T{=}1.0$) \\[2pt]
   $z_k \sim \mathrm{AV}(\cdot \mid h_\ell[q_k])$ \\[2pt]
   $\to$ 4 text explanations per prompt: \\[1.8cm]}; 
\node[stepbadge, left=0.35cm of s3] {3};
 
\path (s3.south) ++(0, 0.9) coordinate (z_center);
\node[zbox] (z2) at ($(z_center) + (-0.575, 0)$) {$z_2$};
\node[zbox] (z1) [left=0.2cm of z2] {$z_1$};
\node[zbox] (z3) [right=0.2cm of z2] {$z_3$};
\node[zbox] (z4) [right=0.2cm of z3] {$z_4$};

\draw[goldline, thick, -{Stealth[length=2mm]}] ($(z1.north) + (0, 0.35)$) -- (z1.north);
\draw[goldline, thick, -{Stealth[length=2mm]}] ($(z2.north) + (0, 0.35)$) -- (z2.north);
\draw[goldline, thick, -{Stealth[length=2mm]}] ($(z3.north) + (0, 0.35)$) -- (z3.north);
\draw[goldline, thick, -{Stealth[length=2mm]}] ($(z4.north) + (0, 0.35)$) -- (z4.north);
 
\node[stepbox, below=1.1cm of s3, minimum height=1.9cm] (s4)
  {Claude Sonnet receives $(z_1,z_2,z_3,z_4)$ \\
   (structured prompt, not free-form summary) \\[2pt]
   $\to$ JSON: detected category, quartile of \\
   first appearance, verbatim quote};
\node[stepbadge, left=0.35cm of s4] {4};
 
\draw[arrowflow] (s1.south) -- (s2.north);
\draw[arrowflow] (s2.south) -- (s3.north);
\draw[arrowflow] (s3.south) -- (s4.north);
 
\node[draw=darktext, thick, fill=graybox, rounded corners=2pt,
      minimum width=8.6cm, minimum height=0.85cm, font=\sffamily\bfseries,
      text=darktext, below=0.7cm of s4] (total)
  {30 prompts $\times$ 4 quartiles $=$ 120 total AV calls};
 
\draw[arrowflow] (s4.south) -- (total.north);
\end{tikzpicture}%
}
\vspace{0.3cm}
\caption{\textbf{Per-prompt pipeline.} Each of the 30 prompts passes through the four stages; the resulting vectors $h_\ell[q_k]$ and explanations $z_k$ are indexed by quartile to strictly preserve the positional signal.}
\label{fig:pipeline}
\end{figure}

\FloatBarrier
\subsection{Cross-Lingual Divergence in Internal Representations}
The analysis of the model's latent thoughts reveals a marked divergence in the contextualization of neutral prompts depending on the language. This variation introduces geographic and migratory biases into the representation space that are not explicit in the original input. Below, three representative cases illustrating this behavior are detailed:

\begin{description}

    \item[Case A01: Health System and Employment] \hfill \\
    \textbf{Original context:} Query about enrollment in EPS (public health insurance) and the Colombian health system when starting a new job.\\
    \textbf{Latent representation:} In Spanish, the model assumes the user is a ``foreigner or newcomer,'' injecting a migratory bias absent in the prompt. In English, the search space shifts towards health insurance in Canada, Germany, or Turkey, linking it to residency procedures.\\
    \textbf{Analysis:} A strong implicit association between employment/health and migration is evident. Processing in Spanish projects local migratory vulnerability, while English completely internationalizes the query, losing the original geographic relevance.

    \item[Case A03: Gastronomy and Cultural References] \hfill \\
    \textbf{Original context:} Query about a traditional Colombian dish (\textit{ajiaco}).\\
    \textbf{Latent representation:} In Spanish, the model recognizes the local traditional gastronomic context, although it generalizes by mixing it with other dishes (\textit{sancocho}, \textit{arepas}). In English, \textit{ajiaco} undergoes a drastic shift towards ``Peruvian, Mexican food'' and invents concepts like ``Colombian paella.''\\
    \textbf{Analysis:} This demonstrates a clear homogenization in English, where Latin American cultural identities are mixed and become interchangeable within the model's latent space.

    \item[Case A05: Education and Financing] \hfill \\
    \textbf{Original context:} Query about how a university educational loan (\textit{Icetex}) works.\\
    \textbf{Latent representation:} In Spanish, the model tends to transform the concept of a loan into a state ``general aid or scholarship.'' In English, the model deflects the query toward international scholarships, mentioning the Erasmus program and French universities.\\
    \textbf{Analysis:} This reflects a severe socioeconomic divergence: processing in Spanish associates the loan with basic local assistance and subsidies, whereas English associates it with international academic mobility.

    \item[Case B01: Institutional Normalization (EPS)] \hfill \\
    \textbf{Original context:} Query including the term ``EPS'' (a Colombian health insurance entity).\\
    \textbf{Latent representation:} In the English version, the model substitutes the term with US insurance companies like Blue Cross and Aetna. In Spanish, the health context is also unrecognized, triggering activations related to unrelated global entities such as Netflix and SAP.\\
    \textbf{Analysis:} The model loses fidelity to the original context, exhibiting a strong tendency to normalize specific local references by replacing them with globalized elements or US-centric entities prevalent in its training data.

    \item[Case B02: Geographic Anchoring (TransMilenio)] \hfill \\
    \textbf{Original context:} Query referencing ``TransMilenio'' (Bogotá's mass transit system).\\
    \textbf{Latent representation:} The model explicitly recognizes it as a Colombian transport system in both languages initially, even when the country is not explicitly mentioned. However, in later processing stages, the signal dilutes into generic urban transport concepts, referencing metros in Madrid, Barcelona, and Tokyo.\\
    \textbf{Analysis:} Highly distinctive geographic entities successfully preserve their national identity and trigger accurate latent representations early on. Yet, this specificity fades as the model shifts toward generic global urban frameworks before final text generation.

    \item[Case B03: Legal Terminology (\textit{Tutela})] \hfill \\
    \textbf{Original context:} Query involving the legal term \textit{``tutela''} (a specific Colombian constitutional protection mechanism).\\
    \textbf{Latent representation:} The juridical signal successfully activates references to the Colombian context only in the Spanish version. In English, this association weakens significantly, and the concept transforms into generic categories like ``human rights petition'' or ``constitutional claim,'' even shifting the context toward Spain.\\
    \textbf{Analysis:} The preservation of specific legal concepts is highly language-dependent. Translation causes the Colombian specificity to dilute into broader international frameworks, demonstrating a semantic normalization in the English latent space.

        \item[Case C01: Health and Employment] \hfill \\
    \textbf{Neutral prompt:} \textit{How can I enroll in private health insurance if I have just started working?} (ES/EN variants).\\
    \textbf{Latent representation:} In Spanish, the model internally generates premises such as \textit{``If you are starting to work in Spain...''} and anticipates terms like \textit{``migratory''} or \textit{``from the United States''}. In English, the model mentions Canada superficially, without linking it to a transitional status.\\
    \textbf{Analysis:} There is a strong implicit association in Spanish between labor/health insertion and emigration, biasing the interpretation towards a context of migratory vulnerability.

    \item[Case C02: Urban Transportation] \hfill \\
    \textbf{Neutral prompt:} \textit{What time does the last metro train run on weekends?}\\
    \textbf{Latent representation:} Processing in Spanish forces an immediate geographic anchoring, generating the anticipation: \textit{``What time is the metro in Barcelona?''}. In English, the internal context is more evenly distributed among various global metropolises (New York, London, Tokyo).\\
    \textbf{Analysis:} This demonstrates a Hispanic-centric localization bias that reduces the model's spatial generalization when faced with generic urban queries in Spanish.

    \item[Case C03: Cultural References] \hfill \\
    \textbf{Neutral prompt:} \textit{What is the traditional recipe for a vegetable soup served in many cultures?}\\
    \textbf{Latent representation:} In Spanish, internal activations are directed towards Ibero-American references, anticipating \textit{``Mexico''} or \textit{``paella''}. In contrast, in English, the search space is oriented towards the Northern Hemisphere, mentioning Italian and Turkish food, and holidays like \textit{Thanksgiving}.\\
    \textbf{Analysis:} A cultural preconditioning is evident in the pre-generation layers, where the language restricts the scope of what the model considers a ``generic culture''.

\end{description}

\end{document}